\documentclass[10pt,journal,compsoc]{IEEEtran}
\usepackage{tabularx}
\usepackage{graphicx}
\usepackage{multirow}
\usepackage{booktabs}
\usepackage{adjustbox}
\usepackage{xcolor}
\usepackage{threeparttable} 
\usepackage{amsmath}

\PassOptionsToPackage{dvipsnames}{xcolor}

\newcommand{\ie}{i.\,e.,\ }

\begin{document}
%
\title{Bare Demo of IEEEtran.cls\\ for IEEE Journals}
\title{Are Large Language Models Emotional Enough? Shifting Emotional Modelling from Specific to General Ones}
\title{Rethink Emotional Models: \\Shifting From Specifically Small Ones to Generally Large Ones}
\title{Rethink Emotion Recognition Modelling: \\Shifting to Generalised Large Models}
\title{New Opportunities for Emotion Recognition in the Large Model Era}
\title{Unveiling New Frontiers in Emotion Recognition Through the Surge of Large-Scale Models}
\title{Revolutionising Emotion Recognition Modelling:\\ A Transition to the Era of Generalised Large Models}
\title{Evolving Perspectives in Emotion Recognition Modelling: The Advent of Generalised Large Models}
\title{Refashioning Emotion Recognition Modelling: The Advent of Generalised Large Models}
%
%
%

\author{Zixing~Zhang,~\IEEEmembership{Senior Member,~IEEE,}
        Liyizhe~Peng,~Tao~Pang,
        Jing~Han,~\IEEEmembership{Senior Member,~IEEE},
        Huan~Zhao,
        Bj\"orn W.\ Schuller,~\IEEEmembership{Fellow,~IEEE}
\thanks{Corresponding authors: Jing~Han (jh2298@cam.ac.uk)} 
\thanks{Z.~Zhang, L.~Peng, T.~Pang, and H.~Zhao are with the College of Computer Science and Electronic Engineering, Hunan University, Changsha, 410082, China. \{zixingzhang, pengliyizhe, pang\_tao, hzhao\}@hnu.edu.cn}
\thanks{J.~Han is with the Department of Computer Science and Technology, University of Cambridge, 15 JJ Thomson Ave, Cambridge CB3 0FD, UK. jh2298@cam.ac.uk}
\thanks{B.\,W.~Schuller is with GLAM -- the Group on Language, Audio, \& Music, Imperial College London, 180 Queen's Gate, Huxley Bldg., London SW7 2AZ, UK, and also with the Chair of Embedded Intelligence for Health Care and Wellbeing, University of Augsburg, Eichleitnerstr.~30, Augsburg 86159, Germany. schuller@ieee.org}
}

%
%

{}


\IEEEtitleabstractindextext{%
\begin{abstract}
After the inception of emotion recognition or affective computing, it has increasingly become an active research topic due to its broad applications. Over the past couple of decades,  emotion recognition models have gradually migrated from statistically shallow models to neural network-based deep models, which can significantly boost the performance of emotion recognition models and consistently achieve the best results on different benchmarks. Therefore, in recent years, deep models have  always been considered the first option for emotion recognition. However, the debut of large language models (LLMs), such as ChatGPT, has remarkably astonished the world due to their emerged capabilities of zero/few-shot learning, in-context learning, chain-of-thought, and others that are never shown in previous deep models. In the present paper, we comprehensively investigate how the LLMs perform in emotion recognition in terms of diverse aspects, including in-context learning, few-short learning, accuracy, generalisation, and explanation. Moreover, we offer some insights and pose other potential challenges, hoping to ignite broader discussions about enhancing emotion recognition in the new era of advanced and generalised large models.
\end{abstract}

}

%
\maketitle
\IEEEpeerreviewmaketitle


\section{Introduction} 

Affect modelling is key for human-centred machine intelligence systems,  allowing for more empathetic, adaptive, and engaging interactions, by considering users' emotions. 
Developing valid and reliable affective models that accurately capture and understand human emotions can pave the way for various applications, including affective dialogue systems, affect-aware recommender systems, and emotion-based adaptive interfaces.  
Over the past decades, the availability of large-scale affective datasets and rich computational power, as well as the advancements in deep learning algorithms, have revolutionised the field of affect modelling research. 

In particular, deep learning techniques, such as neural networks, have proven to be highly effective in capturing complex patterns and relationships within affective data~\cite{latif2021survey, wang2022systematic, jung2019utilizing}.
For instance, Convolutional Neural Networks (CNNs) are efficient to learn and extract relevant features, and to capture local patterns and dependencies, which have demonstrated remarkable success in emotion recognition tasks across various domains and datasets~\cite{DBLP:conf/emnlp/GhosalMPCG19, kollias2020exploiting}.
Likewise, Recurrent Neural Networks (RNNs) are also widely used in affective modelling when processing sequential data like speech and text, by capturing temporal dependencies and contextual information~\cite{zhang2016facing, DBLP:conf/aaai/MajumderPHMGC19}.
Moreover, researchers have also made massive efforts to develop and implement various algorithms and approaches, aiming at more accurate and robust emotion recognition systems to understand human emotions. 
Such efforts include but are not limited to integrating information from different modalities~\cite{han2019emobed, poria2017review}, developing large-scale emotion datasets for model training and validation~\cite{kossaifi2019sewa, bagher-zadeh-etal-2018-multimodal}, transferring knowledge of other relevant tasks by leveraging pre-trained models~\cite{sharma2022multi,DBLP:journals/access/LuTM23}, and increasing the interpretability and providing insights into the decisions~\cite{perikos2021sentiment, kumar2021towards}.

However, all aforementioned works are built typically for emotion recognition tasks, and normally the employed models focus on limited and specific domains. In other words, though prevalent in various emotion-aware systems and applications today, these specialised models lack broad generalisation and adaptability. For instance, a system specifically designed to predict a set of six emotions may not be applicable or effective in predicting an unseen or different emotion. Similarly, an emotion recognition model designed specifically for the Chinese language cannot be directly applied to the English language. These models and systems designed for specific tasks are often referred to as Artificial Narrow Intelligence (ANI). 

In contrast to ANI, a more advanced and comprehensive form of artificial intelligence is Artificial General Intelligence (AGI), being able to learn, adapt, and apply knowledge across a wide range of tasks. While AGI systems may not be specifically tailored for emotion recognition tasks, they have the potential to achieve comparable levels of accuracy and effectiveness as ANI ones that are specifically designed for the task.
In particular, recent breakthroughs in large language models (LLMs), such as GPT-4, are demonstrating significant advancements in natural language processing. These advancements may also open a new frontier for emotion recognition. These models are normally pre-trained over large-scale corpora, and have showcased their strong capabilities in various domains and tasks, such as text generation and natural language understanding.

When utilising LLMs for emotion recognition, there are several potential benefits that may enhance the performance and capabilities of the models: 1) Large Training Data: being trained on vast amounts of data, LLMs can capture more diverse patterns, linguistic cues, and contextual information related to emotions, and thus might lead to better recognition performance; 2) Explanations: LLMs can potentially explain the reasoning behind their decisions, and therefore can increase the interpretability and transparency of the emotion recognition process; 3) Generalisation: LLMs are designed to learn general language patterns and can generalise well to unseen data. This allows them to recognise emotions that may not have been explicitly encountered during training; 4) Cross-Domain Application: LLMs have the potential to be applied across various domains, since they are trained on a wide range of data sources, and thus 
hold the potential to understand emotions expressed in various domains ranging from customer reviews to conversational data, which allows for wider applicability.

Therefore, the focus of this work is on the analysis of the emotion recognition capabilities of LLMs and their implications. Also, we share our views of a future in LLMs-based affective modelling. For this aim, we choose three LLMs and conduct a comparative analysis of their emotion recognition performance against other state-of-the-art (SOTA) non-LLM-based works using seven emotional datasets. This comparison allows us to assess the advancements and effectiveness of LLM-based approaches over specific deep-learning models in the context of emotion recognition.
More specifically, we investigate three different scenarios for LLM-based emotion recognition: 1) emotion prediction of a given sample in the absence of any contextual information; 2) emotion prediction of a sample with context information of that sample; 3) emotion prediction of a sample while an appropriate number of examples are provided as demonstrations. By exploring these three scenarios, we aim to assess LLM-based emotion recognition models' effectiveness in different contextual settings. 
Besides, this work further delves into examining the
generalisation, and interpretability of LLM-based affective modelling. 
Last, we engage in comprehensive discussions on the associated challenges and opportunities, hoping to pave the way for advancements and improvements in the field.

The main contributions of this paper are as follows:
\begin{itemize}
    \item We for the first time carry out a deep detailed investigation of the emotion recognition capability of LLMs. In specific, we systematically evaluated three selected LLMs for emotion recognition tasks on seven datasets from two different languages.
    \item We extensively compare the performance with other state-of-the-art models, showing that LLMs can achieve comparable or superior performance on emotion recognition tasks.
    \item We comprehensively discuss the obtained results and point out challenges and opportunities that are still faced in the research and development of LLM-based affective modelling
\end{itemize}

The structure of this work is organised as follows: Section 2 presents related works; Section 3 introduces the selected LLMs, datasets for evaluation, and implementation details. Experimental results, along with their analysis and comparison, are provided in Section 4. Challenges and opportunities of LLMs in the context of affective modelling are discussed in Section 5. The work concludes in Section 6.

\section{Background}
This section provides a brief overview of the relevant background and concepts.

\subsection{Recent Advances of Emotion Recognition}



Over the past decades, particularly with the emergence of deep learning, significant endeavours have been dedicated to developing effective and robust emotion recognition models. In the following, we briefly introduce related studies primarily focusing on three key aspects/challenges: enhancing accuracy, 
facilitating generalisation, and enabling explainability of emotion recognition models.

One main research direction is to improve the \textit{correctness} of the model, namely increasing the accuracy and precision of emotion recognition models. For this aim, several research studies have investigated various advanced feature extraction techniques, aiming to capture salient emotional representations. The features could be extracted from different modalities, such as speech, video, text, and physiological signals. For instance, three novel domain-specific audiovisual tasks were designed in~\cite{goncalves2022improving} to learn better representations, by exploring large volumes of unlabelled data via self-supervised learning.
There are also a huge number of studies that leverage different deep learning architectures to refine emotion recognition performance. For instance, a Transformer-based model was proposed in~\cite{vazquez2022transformer} to process ECG signals for emotion recognition, and the obtained contextualised representations achieved state-of-the-art performance on the AMIGOS dataset. 
In addition, some other works focus on combining information from multiple modalities, so that a more comprehensive and holistic understanding of users' emotional states can be obtained~\cite{poria2017review}.


Aiming at improving the \textit{generalisation} of emotion recognition models, several advanced deep learning techniques have been explored, such as knowledge transfer~\cite{10096757}, multitask learning~\cite{9947296}, self-supervised learning~\cite{sarkar2020self}, and domain adaptation~\cite{deng2014autoencoder}. Applying these techniques appropriately could make the learnt emotion recognition model more effective across different domains, different contexts, and diverse scenarios. For example, a pre-trained model's knowledge gained from a large amount of data can be transferred to downstream tasks. Specifically, pre-trained language models (PLMs), such as bidirectional encoder representations from transformers (BERT) and their variants have been utilised in text-based emotion recognition and achieved promising performance. In~\cite{adoma2020comparative}, the efficacy of BERT and its variants has been comparatively examined on a cross-culture emotional database and Robustly optimised BERT pre-training Approach (RoBERTa) achieved the best performance among others.

Enhancing \textit{interpretability} is also critical for a reliable affective modelling system, where explanations should be provided along with its predictions. For this reason, advanced techniques have been applied in emotion recognition to improve the model interpretability. Such techniques include model-agnostic interpretation methods~\cite{9651357}, attention mechanisms~\cite{xie2020assisted}, and Bayesian neural networks~\cite{wang2023multimodal}.

These aforementioned efforts facilitate the development of correct, generalisible, and explainable affective computing, and tremendous progress in this research field has been made thanks to these efforts. However, these efforts deal with the three above-mentioned challenges separately, and it is not straightforward to integrate all as a whole.

\subsection{The New Era of Large Models} 
While transfer learning and PLMs have already shown promising advances on various downstream tasks, a new era of large models has started when researchers observed that scaling PLMs can largely lead to further improvement of the model capacity~\cite{bommasani2021opportunities}. 
These large-sized PLMs are termed Large Language Models (LLMs). Compared with smaller PLMs, LLMs significantly extend the model size and data size, even through the model architectures and the pre-training tasks are kept the same~\cite{zhao2023survey}.

Various tech companies and research labs are striving to develop their own LLMs, and among these, there are some popular ones, including ChatGPT and the latest GPT-4 released by OpenAI, and LLaMA models introduced by Meta AI. Particularly, the launch of ChatGPT represents a significant step forward in the evolution of LLMs. Most LLMs are trained in two stages; LLMs normally are first pre-trained on a vast amount of text data, and then further adapted via a reinforcement learning from human feedback (RLHF) algorithm~\cite{zhao2023survey}. While still based on standard deep learning and transfer learning, LLMs display a large performance improvement in terms of accuracy and robustness, achieving state-of-the-art results across multiple domains.

Further, LLMs exhibit surprising generalisation capability on unseen tasks, and are capable of solving a diverse range of complex tasks, even without further fine-tuning on each specialised task. 
This is mainly because LLMs have exhibited new emergent capabilities which were not present in smaller-scale PLMs~\cite{wei2022emergent}.
One such new emergent ability is in-context learning (ICL), which was first introduced by GPT-3 in~\cite{brown2020language}. Being fed a prompt composed of task instructions and demonstrations (input-output examples of that task), the model learns to solve this new task and generate the output of an unseen inference-time example, without any further model training or gradient updates.
Other emergent abilities include step-by-step reasoning, instruction following, program execution, and model calibration~\cite{wei2022emergent}.
With these advanced functionalities, LLMs have attracted increasing and widespread attention, leading to a paradigm shift in AI paradigm, shifting from task-specific models to more general-purpose models. 
In other words, the rapid technical progress of LLMs provides a fertile ground for AGI.

\subsection{At the Crossroad}

Emerging as powerful tools to understand and generate human-readable text, LLMs have demonstrated exceptional capabilities in a wide range of NLP tasks. Their impact on many domains is significant, such as question-answering, code generation, and creative writing. Standing at the crossroads of traditional affective modelling and large models, in this paper, we aim to explore LLMs' impact on affective computing, as the effectiveness of LLMs on this task is not fully investigated, yet. 

Having trained on massive amounts of textual data, LLMs offer exciting possibilities for understanding the subtle variations in emotional content.
Besides, though not trained explicitly for emotion recognition like previous task-specific models, the general-purpose LLMs have strong adaptation capabilities. Especially, prompting-based adaptation methods enable human-like interactions with an LLM via very simple prompts, entailing temporary learning during inference without specialised fine-tuning. 

While LLMs hold promising opportunities in this domain, there still lacks a thorough investigation of the effectiveness of LLMs in affective computing. Specifically, there are still several key questions that need to be answered:
\begin{itemize}
    \item How do LLMs perform in emotion recognition tasks, compared with state-of-the-art task-specific models?
    \item Can one LLM be utilised across different affective computing-related tasks without further training or updating?
    \item To what level can LLMs explain the reasons behind the detected emotion category from a given emotional input?
\end{itemize}
Different from~\cite{amin2023will} which focused on evaluating the effectiveness of ChatGPT on three datasets, this paper explores the efficiency and effectiveness of three LLMs on seven emotional datasets from two languages. More importantly, we explore the emergent capability of in-context learning, showing that with few demonstration examples, LLMs perform better. The generalisation and explainability of LLMs for affective computing are also analysed, which is missing in previous works.

\section{Emotion Recognition with Large Language Models}
In this section, we outline our evaluation strategy for LLM-based emotion recognition. We begin by introducing the three LLMs selected for this evaluation. Then, we give an overview of the seven emotional datasets used in the evaluation. Lastly, we detail the implementation process.

\subsection{Large Language Models}
In this work, we investigate the capabilities of LLMs to perform emotion recognition. For this aim, three LLMs are selected, including ChatGPT, Claude, and Bing Chat.

\textbf{ChatGPT}: 
Released by OpenAI in November 2022, ChatGPT represents a significant advancement in the NLP field. When given a prompt, ChatGPT is designed to provide detailed responses. It builds upon the prior GPT-3.5 model, which has 175 billion parameters, but is specifically optimised for dialogues. Notably, ChatGPT underwent fine-tuning via the 
RLHF method to produce more human-like responses.
The RLHF procedure comprises three stages: supervised fine-tuning, reward model training, and reinforcement learning fine-tuning. Through RLHF, ChatGPT's behaviour is better aligned with human values and preferences. Consequently, ChatGPT excels at human-like communication, generating coherent responses and engaging in meaningful dialogues. This innovation paves the way for diverse applications. In our study, we delved into the emotion recognition capabilities of ChatGPT by interacting with the OpenAI API using their official Python bindings.
For our API interactions, we specified the model as gpt-3.5-turbo, which underpins ChatGPT. We also set the temperature to 0.01 to ensure more focused and deterministic outputs.


\textbf{Claude:} 
Claude, launched in March 2023 by Anthropic, is an LLM-based AI Chatbot.
As a strong competitor to ChatGPT, Claude is also capable of a wide variety of conversational and text processing tasks. These include creative writing, Q$\&$A, project planning, and coding. With the aim of creating AI systems that are helpful, honest, and harmless, the RLHF technique was also employed in Claude~\cite{bai2022training}. Further, Claude was trained using a “Constitutional AI” technique, where a model rather than a human determines the rankings/preferences based on a set of underlying principles~\cite{bai2022constitutional}. In this manner, the output generated by Claude can align with human values and goals from inception.
Currently, access to the Claude API is limited, requiring an application to request access. 
However, the Claude app can be integrated into the Slack workspace free of charge without any application.
Users can engage with Claude by mentioning @Claude in a channel or by sending Claude direct messages. 
For our purposes, we leveraged the Python Slack SDK to mimic functions akin to an API. By obtaining the User Token and Claude Bot ID from Slack and embedding them in our Python code, we can interact with Slack via a Python program. This automation not only streamlines our testing process but also significantly reduces our workload.



\textbf{Bing Chat:}
Microsoft announced the next version of its web search engine -- the new Bing in February 2023, augmented with a variety of cutting-edge technologies from both Microsoft and its partner OpenAI.
With this, Microsoft aims at bringing together search, browsing, and chat into one unified experience. Particularly, the new Bing introduces a chat feature, Bing Chat, designed to deliver answers that feel more conversational than typical search engine responses.
Bing Chat is powered by OpenAI's GPT-4 but customised specifically for search. Also, Bing Chat has access to the internet and thus can provide more up-to-date information. This is different from ChatGPT, which was trained on data only up until September 2021. Additionally, in Bing Chat, the response to a user's query is grounded with search results, incorporating footnotes that trace back to original sources for verification.
Although Bing Chat is exclusive to web access and lacks an API service, we leveraged ChatALL to interface with it. ChatALL serves as a client, granting users the ability to converse with several AI chatbots, Bing Chat included. A notable advantage of using ChatALL is the absence of token limitations when engaging with Bing Chat. It is worth mentioning that Bing Chat offers three distinct response formats: more creative, more balanced, and more precise. For the purposes of our study, we opted for the ``more precise'' format, aiming to obtain succinct and direct answers to our emotion-prediction-related queries.

\subsection{Emotion Datasets}
In this part, we introduce the seven datasets employed in our study. A summary of their statistics can be found in Table~\ref{table:datasets}. As these datasets are publicly accessible, they facilitate easy verification of our results.

\begin{itemize}
    \item \textbf {SST}     
    The Stanford Sentiment Treebank is an English corpus with fine-grained sentiment labels of 11,855 single sentences extracted from movie review data~\cite{socher2013recursive}. These sentences were parsed using the Stanford parser, resulting in a total of 215,154 unique phrases from those parsed trees. Those phrases were then annotated by the classic Mechanical Turk for sentiment into 25 different levels. The obtained annotations for each phrase were then used to define fine-grained and binary versions of the task. For the fine-grained task, each sentence is labelled with one of five sentiment classes: negative, somewhat negative, neutral, somewhat positive, and positive; for the binary task, each sentence is labelled as either positive or negative (the neutral class is ignored). In this study, we do not consider phrase-level sentiment analysis. Sentences were split into training (8544), development (1101), and test splits (2210) for the fine-grained classification task. For the binary task, the three sets have 6920, 872, and 1821 sentences, respectively.

    \item \textbf{Friends} 
    Friends is a corpus derived from the Friends TV shows, comprising transcriptions of 1,000 dialogue spoken by the characters in the show across seasons 1 to 9~\cite{chen2018emotionlines}. Each dialogue was regarded as an annotation task on Amazon Mechanical Turk. The goal was to label each of 14,503 utterances within the 1,000 dialogues into one of seven classes: anger, disgust, fear, joy, sadness, surprise, and neutral. The annotators were instructed to consider the context of the entire dialogue while assigning the sentiment labels. The dataset was further divided into three splits, resulting in 10,561/720, 1,178/80, and 2,764/200 utterances/dialogues in the training, development, and test sets,  separately. 

    \item \textbf{Mastodon}
    The Mastodon dataset was built by crawling social media posts from Mastodon, and non-English posts were filtered out automatically~\cite{cerisara2018multi}. The corpus was completely manually anonymised and annotated by two students. While the dataset was originally designed for sentiment recognition and dialogue act recognition, here, we consider the sentiment analysis task only. Each post was labelled as positive, negative, or neutral. All posts were partitioned into two splits: 1,075 posts from 239 dialogues in the training set, and another 1,142 posts from 266 dialogues in the test set.

    \item \textbf{MOSI}
    The Multimodal Opinion-level Sentiment Intensity (MOSI) dataset~\cite{zadeh2016multimodal} is a multimodal dataset for studying sentiment and subjectivity in opinion videos. The dataset consists of 2199 opinion video segments, segmented from 93 videos generated by 89 distinct English-speaking YouTube users. The sentiment intensity annotation was carried out by master workers via Amazon Mechanical Turk. Each opinion segment was annotated as a sentiment spectrum between highly negative and highly positive in the range [-3,3]: strongly positive (labelled as +3), positive (+2), weakly positive (+1), neutral (0), weakly negative (-1), negative (-2), and strongly negative (-3). Besides, transcriptions are provided for each segment. 
   

    \item \textbf{MOSEI}
    CMU Multimodal Opinion Sentiment and Emotion Intensity (CMU-MOSEI) is the largest dataset of sentence-level sentiment analysis and emotion recognition in online videos~\cite{bagher-zadeh-etal-2018-multimodal}. It contains 23,453 
    sentence utterance videos from 1000 English-speaking YouTube speakers. These sentence utterances were randomly chosen from various topics and monologue videos. Similar to MOSI, the dataset was annotated by 3 master judges via the  Amazon Mechanical Turk platform. Each sentence in the MOSEI dataset is annotated for sentiment on a Likert scale from -3 to +3. It is also annotated for six emotions: happiness, sadness, anger, fear, disgust, and surprise. 

    

    \item \textbf{CH-SIMS}
    A Chinese single- and multi-modal sentiment analysis dataset, CH-SIMS, was introduced in~\cite{yu2020ch}. It contains 2281 refined video segments in the wild,  collected from different movies, TV serials, and variety shows. The dataset includes both multimodal and independent unimodal annotations, allowing both unimodal and multimodal sentiment analysis. Each clip has a manual transcription. The dataset was annotated by five students to label each clip into one of three classes: -1 (negative), 0 (neutral), or 1 (positive). Then, the final sentiment annotation falls into one of five categories: negative, weakly negative, neutral, weakly positive, and positive. For experiments, the data was further split into training (1368), validation (456), and test (457) sets. 

    \item \textbf{M$^3$ED}
    A Multi-modal Multi-scene Multi-label Emotional Dialogue (M$^3$ED) dataset was recently proposed in~\cite{DBLP:conf/acl/ZhaoZ0LJW022}. The dataset contains 990 dyadic emotional dialogues, 9082 turns, and 24,449 utterances, derived from 56 different TV series. M$^3$ED is the first multimodal emotional dialogue dataset in Chinese. Similar to Friends, the dataset is annotated on utterance level with seven emotion categories, including Ekman's six basic emotions (happy, surprise, sad, disgust, anger, and fear) and neutral. 
    
\end{itemize}

\begin{table*}
\begin{threeparttable}
\centering
\caption{Detailed information of the selected seven datasets. \#sp., \#dia., \#total (test) utt., \#words/utt., \#classes denote the number of distinct speakers, dialogues, utterances of the whole dataset and its test subset, words per utterance, and emotional classes.}\label{table:datasets}
\begin{tabular}
{p{1.2cm}<{\centering}p{1.2cm}<{\centering}p{1.2cm}<{\centering}p{1cm}<{\centering}p{1.5cm}<{\centering}p{0.6cm}<{\raggedleft}p{0.7cm}<{\raggedleft}p{2.0cm}<{\raggedleft}p{1.0cm}<{\raggedleft}p{3.5cm}<{\centering}}
\toprule
\textbf{dataset} & \textbf{language} & \textbf{modality} & \textbf{dialogue} & \textbf{data source} & \textbf{\#sp.} & \textbf{\#dia.} & \textbf{\#utt. total (test) } & \textbf{\#words/utt.} & \textbf{\#classes} \\
\midrule

SST & English & t & no & movie review & - & - & 11\,855 (2\,210) & - & 5 (negative, somewhat negative, neutral, positive, somewhat positive) \\

Friends & English & t & yes & Friends TV shows & - & 1\,000 & 14\,503 
 (2\,764) & 10.7 & 7 (neutral, joy, sadness, fear, anger, surprise, disgust)\\

Mastodon & English & t & yes & Mastodon & - & 505 & 2\,217 (1\,142) & - & 3 (positive, neutral, negative)\\

MOSI & English  & a, v, t & no & YouTube & 89 & - & 2\,199 (686) & 12.0 & 7 \{-3, -2, -1, 0, 1, 2, 3\}\\

MOSEI & English  & a, v, t & no & YouTube & 1\,000 & - & 23\,453 (4\,662) & - & 7 \{-3, -2, -1, 0, 1, 2, 3\}\\


CH-SIMS & Mandarin & a, v, t & no & movies, TVs, \& shows & 474 & - & 2\,281 (457) & 15.0 & 5 \{-1.0, -0.8\}\{ -0.6, -0.4, -0.2\} \{0.0\} \{0.2, 0.4, 0.6\}\{0.8, 1.0\}\\

M$^3$ED & Mandarin & a, v, t & yes & TV series & 626 & 990  & 24\,449 (4\,201) & 7.4 & 7 (happy, surprise, sad, disgust, anger, fear, neutral)\\

\bottomrule
\end{tabular}
\end{threeparttable}
\end{table*}

\subsection{Implementation Details}
\textbf{Tasks:} We performed sentiment analysis and emotion recognition tasks on the selected seven datasets, as the testbed to evaluate the affective modelling ability of three LLMs. Particularly, we adhered to others' existing works to ensure a fair comparison. 
For the SST dataset, we undertook both binary and five-class classification tasks. Binary sentiment classification tasks, distinguishing between positive and negative sentiments, were applied to the MOSI, MOSEI, and CH-SIMS datasets. The Mastodon dataset and the MOSI underwent a three-class classification: positive, neutral, and negative. Lastly, for the Friends and M$^3$ED datasets, we executed a seven-class emotion classification.

\textbf{Usage of LLMs:}
When working with LLMs such as ChatGPT, we use a ``prompt'' -- a specific input text -- to elicit a response from the model. In our study on emotion recognition, we thoroughly investigated various prompting strategies. For context-free datasets (SST, MOSI, MOSEI, and CH-SIMS), we assessed two strategies: context-free zero-shot and context-free few-shot prompting. Meanwhile, for context-dependent datasets (Friends, Mastodon, and M$^3$ED), we explored three strategies: context-free zero-shot prompting, context-aware zero-shot prompting, and context-aware few-shot prompting. 
Essentially, `context-free' refers to predicting the emotion-based solely on the content of the sentence itself. In contrast, `context-aware' entails the prediction process that takes into account the surrounding context within the same conversation. Furthermore, `zero-shot' indicates that no prior knowledge or examples are provided for the specific task, whereas `few-shot' implies that a limited number of demonstration examples are given to facilitate inference-time learning.
Detailed prompts for each strategy are provided as follows:
\begin{itemize}
    \item \textbf{context-free zero-shot prompting} (\ie \textbf{w/o context, w/o ICL)}. Classify the sentiment of the sentence to Emotion 1, Emotion 2, ... or Emotion k. No need for you to explain. Don't repeat my sentence. Give me the simplest answer with a list and corresponding number: (provide multiple sentences from a test set with serial number)
    \item \textbf{context-free few-shot prompting} (\ie \textbf{w/o context, w/ ICL)}. Examples of sentiment classification: (provide few training data and corresponding labels from a train set). According to the above examples, classify the sentiment of the sentence as Emotion 1, Emotion 2, ... or Emotion k. No need for you to explain. Don't repeat my sentence. Give me the simplest answer with a list and corresponding number: (provide multiple sentences from a test set with serial number)
    \item \textbf{context-aware zero-shot prompting} (\ie \textbf{w/ context, w/o ICL)}. According to the context of a conversation, classify the sentiment of every sentence to Emotion 1, Emotion 2, ... or Emotion k. No need for you to explain. Don't repeat my sentence. Give me the simplest answer with a list and corresponding number: (provide all sentences of one dialogue with serial number)
    \item \textbf{context-aware few-shot prompting} (\ie \textbf{w/ context, w/ ICL)}. Examples of sentiment classification: (provide few training data and corresponding labels from a train set). According to the above examples, based on the context of conversation, classify the sentiment of every sentence to Emotion 1, Emotion 2, ... or Emotion k. No need for you to explain. Don't repeat my sentence. Give me the simplest answer with a list and corresponding number: (provide all sentences of one dialogue with serial number) 
\end{itemize}
Note that, $K$ is the number of sentiment/emotion categories, determined by each specific task. 

For example, the prompt is ``Classify the sentiment of sentence to Positive or Negative'' for MOSEI as $k=2$. The instruction ``No need for you to explain. Don't repeat my sentence. Give me the simplest answer with a list and corresponding number.'' is aimed at a more concise and clearer response, making it easier to retrieve the desired sentiment/emotion prediction. 
Furthermore, for those binary datasets (positive and negative), we add ``Don’t answer neutral.'' into the prompt, which can significantly reduce the number of neutral responses. However, there are still unexpected neutral responses in the binary classification (positive vs negative) in such a manner. 
In addition, for any sentence, whether from the training set (as task demonstrations) or from the test set, we add two asterisks (**) as separators before and after the sentence. These separators can make it easier for the model to distinguish the beginning and end of sentences, avoiding the model considering multiple sentences as one paragraph and only providing one label instead.
Also, when performing zero/few-shot prompting on context-free datasets, incorporating multiple sentences into a single query notably decreased the evaluation's time and cost requirements. Specifically, 100 test sentences were fed in one prompt for zero-shot prompting, while 50 sentences in one prompt for few-shot prompting.
In contrast, when performing zero-shot prompting on context-dependent datasets, we reduced the number of sentences to one. 
This was to prevent the model from gaining any contextual information.
Last, the number of task demonstrations added to the prompt under few-shot prompting settings cannot exceed 100 due to the limited number of tokens (4096) set in ChatGPT. We selected an equal number of sentences per sentiment/emotion category: 50 sentences per category for binary classification, 30 per category for three-class classification, 20 per category for five-class classification, and 10 per category for seven-class classification.
To ensure fairness in comparison between the three LLMs, for ICL strategies, we applied the same training data points per dataset as task demonstrations when testing each LLM.

After collecting the LLM responses, we post-processed them to address any inconsistencies. Notably, due to occasional model instability, we sometimes received results outside our predefined classification range. For instance, the model might return ``Embarrassed" or ``Grateful", which are not within our specified emotion categories. To counteract this, we adjusted the model's temperature setting it to its maximum value of 2 and re-prompted the model until we obtained a result within the expected range. If, after five attempts, we still did not get a response within our classification range, we deemed the sentiment classification for that sentence as incorrect and randomly assigned it to an incorrect classification category.


\textbf{Model Explainability of LLMs:} Providing clear and understandable explanations for the decisions made by complex models is an active area of research. 
As LLMs generate natural language responses, one potential approach could be to design prompts that request the models to elucidate the reasoning behind their decisions in human-readable format. To showcase the potential of this approach, we opted to conduct an exploratory experiment, using ChatGPT and selecting a few conversations from the Friends dataset.
For each conversation, one of the sentences was chosen to be asked for the explanation behind the model's prediction. To solicit this additional information, we appended the following request to the original prompt this further request ``In addition, please explain how you judge the emotion of the $<$order of the selected sentence$>$ sentence in the conversation based on the context".
The primary aim of this experiment was to showcase whether ChatGPT is capable of providing sound explanations for its emotion recognition decisions. By doing so, we hoped to gain a deeper insight into the model's decision-making process and to further explore the possibilities of integrating XAI principles with LLM capabilities.

\textbf{SOTA specialised models:} For performance comparison with specialised affective computing models, we selected recently published SOTA works with competitive performance on each selected dataset respectively.

\textbf{Evaluation Metrics:} 
The primary metrics we employed for performance evaluation were accuracy and the macro F1 score. For the M$^3$ED dataset, we used the weighted average F1 score to ensure a fair comparison with other studies utilising the same dataset. Similarly, for the Mastodon dataset, we calculated both macro precision and macro recall.

\section{Evaluation and Discussions}
We conducted experiments using seven datasets, with the comparative results presented in Tables 2-5. Compared with SOTAs benchmarks, the LLM-based model demonstrates comparable, if not superior, performance.

\begin{table*}
\centering
\caption{Performance comparison between LLMs and a SOTA study on the {\bf Mastodon} dataset measured by accuracy (Acc), macro-F1 (F1), macro-precision (Precision), and macro-recall (Recall).}
\begin{tabular}
{p{5.5cm}<{\centering}p{1.5cm}<{\centering}p{1.5cm}<{\centering}p{1.5cm}<{\centering}p{1.5cm}<{\centering}}
\toprule 
 Model [\%] & Acc & F1 & Precision & Recall\\
\midrule 
 DARER (2022)~\cite{DBLP:conf/acl/XingT22} & - & 59.59 & 56.04 & 63.33\\
 \midrule 
 ChatGPT (w/o context) & 65.50 & 64.46 & \textbf{72.04} & 63.39\\
 ChatGPT (w/ context) & 65.41 & 65.57 & 67.20 & 65.11\\
 ChatGPT (w/ context + w/ ICL) & \textbf{65.94} & 66.09 & 67.72 & 65.48\\
 Claude (w/o context) & 63.05 & \textbf{66.28} & 66.33 & 66.88\\
 Claude (w/ context) & 64.02 & 63.25 & 63.85 & 66.23\\
 Claude (w/ context + w/ ICL) & 64.27 & 64.51 & 64.75 & \textbf{66.97}\\
 Bing Chat (w/o context) & 60.51 & 60.77 & 62.17 & 62.94\\
 Bing Chat (w/ context) & 60.95 & 61.19 & 61.99 & 63.28\\
 Bing Chat (w/ context + w/ ICL) & 63.66 & 63.94 & 64.14 & 65.42\\
\bottomrule 
\end{tabular}
\label{tab:mastodon}
\end{table*}

\begin{table*}
\centering
\caption{Performance comparison on {\bf Friends} (first half) and {\bf M$^3$ED} (second half) in terms of accuracy (Acc), F1, and unweighted accuracy (UA). The best results are in bold. Note that, F1 indicates macro-F1 and weighted average F1 for Friends and M$^3$ED, respectively. The ratio of each emotion class in the test set is given under each emotion class.}
\begin{tabular}
{p{4.5cm}<{\centering}p{0.7cm}<{\centering}p{0.7cm}<{\centering}p{0.7cm}<{\centering}|p{0.7cm}<{\centering}p{0.7cm}<{\centering}p{0.7cm}<{\centering}p{0.7cm}<{\centering}p{0.7cm}<{\centering}p{0.7cm}<{\centering}p{0.7cm}<{\centering}}
\toprule 
 {Friends} & Acc & F1 & UA & Neu & Joy & Sad & Fea & Ang & Sur & Dis\\
 Model [\%] & & & & (57.89) & (13.68) & (3.82) & (1.44) & (7.24) & (12.87) & (3.06)\\
\midrule 
 CNN-BiLSTM (2017)~\cite{DBLP:conf/acl/PoriaCHMZM17}& 77.40 & - & 39.40 & \textbf{87.00} & 60.30 & 28.70 & 0.00 & 32.40 & 40.90 & 26.70\\
 BERT+SRL-GNN-8 (2020)~\cite{DBLP:conf/mm/HeatonS20}& 72.10 & - & 53.71 & 84.32 & 69.92 & 48.39 & 31.03 & 47.06 & \textbf{73.51} & 21.74\\
 XLNet+SRL-GNN-8 (2020)~\cite{DBLP:conf/mm/HeatonS20}& 72.82 & - & 53.41 & 85.34 & 73.98 & \textbf{67.74} & 3.45 & 60.00 & 61.59 & 21.74\\
 PRE-CODE (2020)~\cite{DBLP:conf/emnlp/JiaoLK20}& \textbf{81.30} & \textbf{65.90} & - & - & - & - & - & - & - & - \\
 \midrule 
 ChatGPT (w/o context) & 72.29 & 54.31 & 52.33 & 85.31 & 73.68 & 38.82 & 40.63 & 37.27 & 53.85 & 36.76\\
 ChatGPT (w/ context) & 63.65 & 51.92 & \textbf{59.43} & 69.31 & 77.30 & 61.18 & 68.75 & 47.83 & 34.27 & \textbf{57.35}\\ 
 ChatGPT (w/ context + w/ ICL) & 63.38 & 50.26 & 57.89 & 67.37 & \textbf{83.55} & 51.76 & 62.50 & 54.04 & 36.01 & 50.00\\
 Claude (w/o context) & 56.63 & 44.16 & 52.74 & 58.42 & 65.10 & 47.05 & 59.37 & 64.59 & 43.70 & 30.88\\
 Claude (w/ context) & 51.51 & 41.22 & 56.53 & 49.34 & 76.64 & 61.18 & \textbf{81.25} & \textbf{80.75} & 17.13 & 29.41\\ 
 Claude (w/ context + w/ ICL) & 58.43 & 44.86 & 53.92 & 63.30 & 63.49 & 43.53 & 68.75 & 69.57 & 33.57 & 35.29\\
 Bing Chat (w/o context) & 40.31 & 33.52 & 40.87 & 37.22 & 58.55 & 36.47 & 31.25 & 52.82 & 30.07 & 39.71\\
 Bing Chat (w/ context) & 55.62 & 45.25 & 56.17 & 52.06 & 72.70 & 65.88 & 37.50 & 71.43 & 45.10 & 48.53\\ 
 Bing Chat (w/ context + w/ ICL) & 56.91 & 44.86 & 53.53 & 56.25 & 75.66 & 51.76 & 43.75 & 62.73 & 43.36 & 41.18\\
 \midrule 
 \midrule 
 {M$^3$ED} & Acc & F1 & UA & Neu & Joy & Sad & Fea & Ang & Sur & Dis\\
 Model [\%]& & & & (44.16) & (8.52) & (17.47) & (1.55) & (17.52) & (5.59) & (5.19)\\
 \midrule 
 DialogueGCN (2019)~\cite{DBLP:conf/emnlp/GhosalMPCG19}& - & 46.09 & - & - & - & - & - & - & - & -\\
 DialogueRNN (2019)~\cite{DBLP:conf/aaai/MajumderPHMGC19}& - & 48.80 & - & - & - & - & - & - & - & -\\
 MDI (2022)~\cite{DBLP:conf/acl/ZhaoZ0LJW022}& - & 49.42 & - & - & - & - & - & - & - & -\\
 \midrule 
 ChatGPT (w/o context) & 44.47 & 40.40 & 31.89 & 69.11 & 39.66 & 19.35 & 24.62 & 24.73 & 25.11 & 20.64\\
 ChatGPT (w/ context) & 45.39 & 43.00 & 35.91 & 65.34 & 35.47 & 27.11 & 29.23 & 26.77 & 34.89 & 32.57\\ 
 ChatGPT (w/ context + w/ ICL) & 46.32 & 45.39 & 32.33 & 76.06 & 29.61 & 13.90 & 33.85 & 27.58 & 19.15 & 26.15\\
 Claude (w/o context) & 34.90 & 34.83 & 31.10 & 33.32 & 19.27 & 20.16 & 36.92 & \textbf{70.52} & 37.02 & 0.46\\
 Claude (w/ context) & 53.73 & 50.14 & 34.14 & \textbf{83.94} & 24.58 & 24.66 & 29.23 & 46.74 & 17.02 & 12.84\\ 
 Claude (w/ context + w/ ICL) & \textbf{53.80} & \textbf{52.83} & 42.56  & 75.20 & 29.33 & 32.02 & \textbf{52.31} & 47.42 & 44.68 & 16.97\\
 Bing Chat (w/o context) & 36.44 & 38.51 & 36.79 & 37.90 & 47.49 & 22.21 & 23.08 & 37.23 & 61.13 & 27.52\\
 Bing Chat (w/ context) & 43.42 & 45.36 & 43.59 & 45.39 & \textbf{54.19} & 27.38 & 35.38 & 46.47 & \textbf{70.64} & 25.69\\ 
 Bing Chat (w/ context + w/ ICL) & 47.73 & 50.53 & \textbf{48.22} & 52.78 & 48.04 & \textbf{32.43} & 49.23 & 45.65 & 55.74 & \textbf{53.67}\\
\bottomrule 
\end{tabular}
\label{tab:emotions}
\end{table*}

\begin{table*}
\centering
\caption{Performance comparison between LLMs and SOTA works on the {\bf MOSI} and {\bf MOSEI} datasets measured by accuracy (Acc) and macro-F1 (F1).}
\begin{tabular}{c c c | c c | c c}
\toprule
\multicolumn{1}{c}{} & \multicolumn{2}{c}{MOSI-2} & \multicolumn{2}{c}{MOSI-3} & \multicolumn{2}{c}{MOSEI} \\
\cmidrule(lr){2-3} \cmidrule(lr){4-5} \cmidrule(l){6-7}
Model [\%] & Acc & F1 & Acc & F1 & Acc & F1 \\
\midrule
TFR-Net (2021)~\cite{DBLP:conf/mm/YuanLXY21} & 83.49 & - & - & - & - & - \\
CHFN (2022)~\cite{DBLP:conf/mm/GuoTDDK22} & 85.20& - & - & - & - & - \\
SeqSeq2Sent (2018)~\cite{DBLP:journals/corr/abs-1807-03915} & - & - & 77.00 & - & -& -\\
CTFN (2021)~\cite{DBLP:conf/acl/TangLJCZK20} & - & - & 80.79 & - & - & -\\
TBJE (2020)~\cite{DBLP:journals/corr/abs-2006-15955} & - & - & - & - & 81.90 & - \\
COGMEN (2022)~\cite{DBLP:conf/naacl/JoshiBJSM22} & - & - & - & - & 84.42 & - \\
\midrule
ChatGPT (w/o ICL) & 86.13 & 85.92 & 73.62 & 62.21 & 85.60 & 84.43 \\
ChatGPT (w/ ICL) & \textbf{89.18} & \textbf{88.93} & 76.38 & 63.32 & 80.74 & 79.80 \\
Claude (w/o ICL) & 87.04 & 86.55 & 79.88 & 63.67 & \textbf{85.83} & \textbf{84.81} \\ 
Claude (w/ ICL) & 88.72 & 88.37 & \textbf{82.65} & \textbf{63.92} & 82.11 & 81.33 \\
Bing Chat (w/o ICL) & 70.73 & 70.72 & 65.60 & 55.76 & 69.84 & 68.36 \\
Bing Chat (w/ ICL) & 88.26 & 88.12 & 67.20 & 55.97 & 72.01 & 70.28 \\
\bottomrule
    \end{tabular}
    \label{tab:mosi}
\end{table*}

\begin{table*}
\centering
\caption{Performance comparison between LLMs and SOTA works on the {\bf SST} and {\bf CH-SIMS} datasets measured by accuracy (Acc) and macro-F1 (F1).}
\begin{tabular}{c c c | c c | c c}
\toprule
\multicolumn{1}{c}{} & \multicolumn{2}{c}{SST-2} & \multicolumn{2}{c}{SST-5} & \multicolumn{2}{c}{CH-SIMS} \\
\cmidrule(lr){2-3} \cmidrule(lr){4-5} \cmidrule(l){6-7}
Model [\%] & Acc & F1 & Acc & F1 & Acc & F1 \\
\midrule
GNN (2020)~\cite{DBLP:conf/acl/ZhangYCWWW20}& 92.13 & - & - & - & - & - \\
BT-TAPT (2021)~\cite{DBLP:journals/corr/abs-2107-10474}& 92.40 & - & - & - & - & - \\
SEMGraph-P (2022)~\cite{DBLP:conf/emnlp/WangLD0X22}& 94.23 & - & - & - & - & - \\
ELECTRA (2020)~\cite{DBLP:conf/emnlp/ClarkLLM20}& - & - & 57.89 & - & - & -\\
SentiLARE (2020)~\cite{DBLP:conf/emnlp/KeJLZH20}& - & - & 58.59 & - & - & -\\
SentiWSP (2022)~\cite{DBLP:conf/emnlp/0007L0LSZGGD22}& - & - & \textbf{59.32} & - & - & -\\
MLF-DNN (2020)~\cite{yu2020ch}& - & - & - & - & 80.26 & -\\
\midrule
ChatGPT (w/o ICL) & 92.32 & 92.32 & 53.36 & 49.43 & 79.66 & 78.78 \\
ChatGPT (w/ ICL) & 95.31 & 95.31 & 54.50 & \textbf{52.41} & 87.28 & 86.43 \\
Claude (w/o ICL) & 95.31 & 95.30 & 49.86 & 40.44 & \textbf{88.70} & \textbf{87.44} \\
Claude (w/ ICL) & \textbf{95.86} & \textbf{95.86} & 51.27 & 46.19 & 77.40 & 76.86 \\
Bing Chat (w/o ICL) & 81.56 & 81.53 & 48.64 & 48.63 & 72.03 & 70.99 \\
Bing Chat (w/ ICL) & 85.64 & 85.64 & 50.32 & 50.25 & 82.77 & 82.09 \\
\bottomrule
\end{tabular}
\label{tab:sst}
\end{table*}

\subsection{In-Context Learning}
Equipping with a few demonstration examples, few-shot prompting can significantly improve the LLM's performance in affective modelling tasks.

As shown in Table~\ref{tab:sst} and Table~\ref{tab:mosi}, on these four context-free sentiment analysis datasets, few-shot prompting surpasses zero-shot prompting in most cases. For example, when testing with ChatGPT, the few-shot prompting strategy obtained increased accuracy (95.31\% over 92.32\%) on  the SST-2 task. For the same task, Claude and Bing Chat also gain benefit from the in-context learning strategy, achieving improved accuracy when compared with zero-shot prompting where demonstration examples are not available. Similar observations can be found on SST-5, CH-SIMS, MOSI-2, MOSI-3, and MOSEI as well. 

Table~\ref{tab:mastodon} presents the result performance on the Mastondon dataset. As this dataset is composed of dialogues, i.\,e., posts/sentences with context information, we consider the performance comparison between zero-shot and few-shot promptings when the whole test dialogue is fed into the model as the input query. Table~\ref{tab:mastodon} shows that few-shot prompting outperforms zero-shot prompting in all measures on this dataset.

For multi-class emotion recognition tasks on two context-dependent datasets (Friends and M$^3$ED), results in terms of WA, UA, F1, and accuracy per emotion category are given in Table~\ref{tab:emotions}. When the performance metrics of WA, UA, and F1 is concerned, performance enhancements are obtained by the in-context learning strategy in 11 out of the total 18 cases (derived from 3 measures * 3 LLMs * 2 datasets = 18 cases).

\subsection{Accuracy}
For each of the selected datasets, SOTA performance from specialised models in previous works is provided for comparison. It is shown that, in general, LLM-based models yield competent or even better performance, both in binary classification and multi-class classification tasks. 
For instance, on the SST-2 and CH-SIMS datasets (cf.\  Table~\ref{tab:sst}), the best performance in accuracy is obtained by Claude, reaching 95.86\% and 88.70\%, respectively, outperforming SOTA specialised models.

When comparing the three selected LLMs, it is found that in general, ChatGPT and Claude are better than Bing Chat. For instance, as shown in Table~\ref{tab:mosi}, on the MOSI-3 and MOSEI datasets, Bing Chat's performance is lower than ChatGPT and Claude. One of the potential reasons for the inferiority of Bing Chat on affective modelling is that the model is tailored for search. However, there is no consistent observation suggesting a particular LLM or prompting strategy as superior over the rest, as the optimal model and strategy setting vary across different datasets.

Further, we observe that incorporating context information (both context within the dialogue and context knowledge from the demonstrations) could enhance the recognition performance for emotional sentences within dialogues. Table~\ref{tab:mastodon} and Table~\ref{tab:emotions} display results on context-available datasets. In all 10 global metrics (4 for Mastodon, 3 for Friends, and 3 for M$^3$ED), 7 of them are improved when ChatGPT is deployed, 8 metrics are increased by Claude, and all 10 metrics are increased when leveraging Bing Chat. 

Not only do LLMs achieve comparable overall performance, but they also display better recognition performance in minority emotion classes. The ratio of each emotion class within two emotion datasets is listed in Table~\ref{tab:emotions}. 
Conventional models tend to struggle with emotions constituting less than 10\% of the dataset samples compared to more prevalent emotion classes. For example, the `fear' emotion class (which makes up 1.44\% of the Friends dataset) achieved a peak accuracy of 31.03\% in state-of-the-art (SOTA) methodologies. In contrast, LLMs across various prompting strategies outperformed this, with Claude achieving the best result at 81.25\%. Similarly, Claude's accuracy of recognising the `angry' emotion in Friends, at 80.75\%, 
was markedly superior to SOTA results. Comparable trends are observed with other minority emotions on the M$^3$ED dataset. This can be attributed to the vast and diverse text LLMs have been trained on, which encompasses a broader emotional range than a specific dataset.




\subsection{Generalisation}
Our experimental results show that, without explicitly training on a specific dataset, LLMs exhibit an impressive generalisation ability in multiple affective modelling tasks. Instead of learning specialised models for each of the seven distinct emotional datasets separately, all three selected models have shown their capabilities in managing varied emotion recognition tasks, underlining their versatility and broad applicability.

In particular, as discussed earlier, the LLMs could achieve comparable performance in zero-shot prompting scenarios. And this performance is further enhanced by the implementation of few-shot prompting, indicating that LLMs can adapt and learn from limited contextual information during the inference. 

Most importantly, the LLMs show a strong generalisation ability in their emotion analysis across different corpora and domains. The datasets utilised in this study encompass a wide array of text types, including TV show scripts, movie reviews, social media posts, and YouTube video transcriptions. Despite the variances in style, tone, and context inherent to these data sources, LLMs consistently produce reliable emotion recognition results.

Additionally, the results attest to the LLMs' cross-language generalisation capability. The corpora used in this study spanned both English and Chinese languages, further emphasising the models' adaptability. Such cross-language generalisation ability holds considerable potential for universal emotion recognition models that overcome language barriers.




\subsection{Explanation}
To further explore the explainability of LLMs in their emotion recognition decisions, we delve into ChatGPT's explanations for two sample conversations, as showcased in Table~\ref{tab:conversation}. This provides an initial insight into how LLMs rationalise their decisions in the context of emotion recognition.

As presented in Table~\ref{tab:conversation}, ChatGPT provides reasonable explanations for the sentences under analysis, effectively utilising the context information within the conversation. For example, taken out of context, the sentence ``Ohh, Well, this is just perfect!'' might seem positive. However, when considering the preceding conversation, the model correctly identifies it as conveying a negative emotion.
It is important to note that while these intelligible explanations offer some insight into the model's decision-making process, the correlation between these human-readable explanations and the actual decision-making mechanism of the model is still ambiguous. This relationship warrants further investigation, which is crucial for establishing more reliable and transparent emotion recognition systems.

\begin{table*}[!t]
\centering
\caption{Two examples illustrating ChatGPT's explanations for conversational emotion recognition. The sentences selected for explanation, along with their corresponding decisions and associated explanations, are highlighted for clarity.}
\begin{tabular}{cp{0.75\textwidth}}
\hline
\textbf{Example 1}& \\
\textbf{Prompt}: & According to the context of the conversation, classify the emotion of every sentence to Neutral, Disgust, Surprise, Anger, Joy, Sadness, or Fear. Give me the simplest answer with a list and corresponding number. In addition, please explain how you judge the emotion of the \colorbox{lightgray}{seventh} sentence in the conversation based on the context:\\
&1.Joey: Hey!\\
&2.Rachel: Hi!\\
&3.Joey: Oh you know uh Kash, really liked you the other day. He said he thought you were charming.\\
&4.Ross: Okay I-I just have to stop by my place first.\\
&5.Joey: To tape the game? You do this every time Ross, you're\\
&6.Ross: No-no, I-I have to see if this apartment became available.\\
&\colorbox{lightgray}{7.Joey: Oh, you're switching apartments?}\\
&8.Ross: It's not for me, it's for Rachel.\\
&9.Joey: But Rachel has an apartment.\\
&10.Ross: Yeah, but when the baby comes she's gonna want to move.\\
&11.Joey: She is?\\
&12.Ross: Yeah, you didn't expect her to live there with a baby did you?\\
&13.Joey: I guess I didn't really think about it.\\
&14.Ross: Ezel! Hey Gunther! You're an ezel!\\
&15.Ross: Damnit!\\
\textbf{ChatGPT response:} & 1. Joy 2. Joy 3. Joy 4. Neutral 5. Anger 6. Fear \colorbox{lightgray}{7. Surprise} 8. Neutral 9. Surprise 10. Neutral 11. Surprise 12. Neutral 13. Neutral 14. Disgust 15. Anger\\
& \colorbox{lightgray}{Explanation for the seventh sentence:} The emotion for the seventh sentence is "Surprise". Joey's response, "Oh, you're switching apartments?", indicates surprise or astonishment in reaction to Ross's statement about looking for a new apartment for Rachel. The word "Oh" at the beginning of the sentence and the questioning tone imply that Joey wasn't expecting this news, and it caught him off guard. This element of surprise is the emotional judgment based on the context of the conversation.\\
\hline
\textbf{Example 2}& \\
\textbf{Prompt}: & According to the context of the conversation, classify the emotion of every sentence to Neutral, Disgust, Surprise, Anger, Joy, Sadness, or Fear. Give me the simplest answer with a list and corresponding number. In addition, please explain how you judge the emotion of the \colorbox{lightgray}{ninth}
sentence in the conversation based on the context:\\
&1.Phoebe: Hey!\\
&2.Monica: Hey.\\
&3.Phoebe: Hey, why are you mopping your ceiling?\\
&4.Monica: Oh, there's banana on it.\\
&5.Rachel: You said she was bald.\\
&6.Phoebe: Yeah, she was bald, she's not now.\\
&7.Rachel: How could you not tell me that she has hair?\\
&8.Phoebe: I don't know, I hardly ever say that about people.\\
&\colorbox{lightgray}{9.Rachel: Ohh, well, this is just perfect!}\\
&10.Phoebe: Well I'm sorry, I thought you said it was okay.\\
&11.Rachel: Yeah, I said what was okay when I thought she was some weird bald chick. I mean, y'know, that girl has hair got\\
&12.Phoebe: Well, maybe it won't work out. Maybe Ross won't like her personality.\\
&13.Rachel: Why, does she have a bad personality?\\
&14.Phoebe: Oh no, Bonnie's the best!\\
\textbf{ChatGPT response:} & 1. Joy 2. Neutral 3. Surprise 4. Disgust 5. Surprise 6. Neutral 7. Anger 8. Neutral \colorbox{lightgray}{9. Anger} 10. Neutral 11. Anger 12. Neutral 13. Neutral 14. Joy \\
&\colorbox{lightgray}{Explanation for the emotion of the ninth sentence:} In the ninth sentence, "Ohh, well, this is just perfect!" Rachel displays anger and frustration. She initially thought Bonnie was bald, and now that she knows Bonnie has hair, Rachel's reaction suggests a negative response. The phrase "this is just perfect" is used sarcastically and negatively, indicating Rachel's annoyance and aligning with the emotion of "Anger."\\
\hline
\hline
\end{tabular}
\label{tab:conversation}
\end{table*}

\section{Challenges and Opportunities}

In this section, we delve into several challenges and opportunities associated with the utilisation of LLMs for affective computing tasks.

\textbf{Robustness}
The robustness of LLMs in affective computing tasks is also a crucial topic. 
When users interact with LLMs like ChatGPT, they may introduce a range of natural errors, encompassing typographical mistakes and grammatical inaccuracies. These errors mirror the imperfect, real-world language usage seen in everyday communications. Hence, it becomes pivotal to investigate the resilience of LLMs to such inconsistencies. It's equally important to explore strategies to enhance model robustness against this ``noise'', especially when discerning emotions from textual content. How well these models handle and respond to these errors not only tests their robustness but also their capability to simulate human-like comprehension and interaction. An LLM's capacity to correctly interpret sentiment or emotional intent in the face of such errors may serve as a strong indicator of its utility and effectiveness in real-world scenarios.

\textbf{Adaptation/fine-tuning}
In this present study, we have explored the capabilities of off-the-shelf LLMs without any additional adaptation/fine-tuning. Although LLMs are trained on vast amounts of text data across various domains, their abilities can be further tailored to specific goals through continual pre-training or fine-tuning by leveraging domain-specific data.

In the context of affective computing, to further enhance the affective modelling ability of LLMs, it would be of interest to deploy further pre-training approaches such as domain-adaptive pretraining (DAPT)~\cite{gururangan-etal-2020-dont} and task-adaptive pretraining (TAPT)~\cite{howard-ruder-2018-universal}. In DAPT, an LLM undergoes additional unsupervised pre-training
on a large corpus of unlabelled domain-specific data, with the aim of adapting the LLM to a particular domain. Conversely, in TAPT, an LLM is further pre-trained on the unlabelled training set for a specific task. Compared to DAPT, TAPT uses a significantly smaller, but far more task-relevant pretraining corpus. Research in~\cite{gururangan-etal-2020-dont} suggests that tailoring a pre-trained model to the domain of a target task via multiphase adaptive pretraining can significantly enhance task performance. It may be valuable to investigate whether further pretraining an LLM towards the domain of the emotion recognition corpus of interest can yield benefits. However, it is worth paying attention to issues such as the time and cost of pre-training, as well as the potential for incurring inductive biases.

In addition, fine-tuning an LLM in a supervised learning way can provide benefits for a specific affective modelling task. For instance, the original model may excel in tasks such as binary sentiment classification, but underperform in more nuanced sentiment analysis tasks such as seven-class sentiment classification. Likewise, the original model may be good at analysing emotions from structured movie reviews, but struggle with less formal and more diverse forms of expression in social media posts. In these conditions, it would be helpful to conduct further fine-tuning for domain-specific adaptation for optimal performance. One primary approach for fine-tuning LLMs is instruction tuning. This approach involves formatting the specific task training data with natural language task descriptions to guide the LLM towards the task objective. Along with the paired input-output training data, these specially designed task descriptions in natural language play a pivotal role in helping the LLM to understand the desired goal. In the realm of affective computing, we may fine-tune an LLM using customised emotion recognition dataset(s) to enhance the model's adaptability. Interestingly, even when using the same labelled instances on the same dataset, different task instructions can lead to varying performance outcomes. This indicates the significant impact that the phrasing and clarity of task descriptions can have on the effectiveness of instruction tuning~\cite{wei2021finetuned}. It could be highly beneficial to investigate directions such as identifying the most effective task descriptions for emotional datasets, exploring strategies to optimise fine-tuning using a minimal amount of target domain emotional data, or how to efficiently fine-tune an LLM for affective computing. 

Finetuning LLMs enables models to learn domain-specific knowledge, using a smaller dataset within the target domain. However, given the massive number of parameters that LLMs possess, full-parameter tuning can be highly resource-intensive, potentially leading to increased computational costs and time. To address this, several parameter-efficient fine-tuning (PEFT) techniques have been devised, which only fine-tune a small number of (extra) parameters and meanwhile maintain the strong performance~\cite{ding2023parameter}. These include strategies such as prefix tuning which optimises trainable vectors added to each Transformer block as a prefix, prompt tuning which learns task-specific prompt embeddings to complement the input text embeddings, adapter tuning which incorporates small neural network modules without changing existing parameters~\cite{houlsby2019parameter}, and low-rank adaptation (LoRA) which decomposes the weight update matrix into lower-dimensional matrices without losing too much important information~\cite{hu2021lora}. It would be interesting to examine the effectiveness of these various PEFT techniques to tailor existing open-source LLMs for emotion recognition tasks. This will expand our understanding of how to most effectively enhance LLMs to meet specific aims in the field of affective computing.


\textbf{Privacy and security} Ensuring user privacy and data security are crucial considerations when deploying LLMs for affective computing in real-world applications. The processed text is inherently sensitive, including users' emotional states, personal experiences, and various forms of identifiable information. Thus, it is essential to implement robust safeguards to protect this information and address privacy concerns when leveraging the advantages of LLMs. As highlighted in~\cite{plant2022you}, an empirical investigation indicates that larger and more complex models are more prone to leaking private information. During their experiments, four evaluated LLMs inadvertently disclosed demographic information (location, age, and gender) to varying degrees during the sentiment analysis process~\cite{plant2022you}. Moreover, they also conducted a critical evaluation of various privacy-preserving technologies, such as differential privacy, to quantify the impact of several privatising methods on alleviating information leakage. An interesting research direction could be the exploration of solutions that effectively balance performance and privacy, with the goal of developing privacy-preserving LLM-based affective computing models.

\textbf{Model compression}
On one hand, LLMs are powerful and show appealing performance across varied tasks. On the other hand,  they also present challenges due to their substantial memory requirements during the inference stage, making deployment costly. In the context of affective computing, computation often takes place locally rather than on cloud-based systems. This approach is primarily adopted to address privacy and security concerns, as it prevents sensitive data from being transmitted elsewhere. However, these local processors are frequently deployed on resource-constrained devices such as mobile phones, where the substantial size and computational requirements of LLMs pose a significant challenge. Therefore, it becomes essential to explore model compression techniques, such as quantisation methods~\cite{gholami2022survey}, to adapt these powerful models for local deployment. These approaches aim to reduce the memory footprint and latency of LLMs and meanwhile maintaining high accuracy, making them more suitable for implementation on local, resource-limited devices. By pursuing this research direction, we can potentially bring the power of LLMs to real-world applications within the area of affective computing, while respecting user privacy and security.

\textbf{Large Multimodal Models:} 
In March 2023, openAI released GPT-4~\cite{OpenAI2023GPT4TR}, extending the text input to multimodal signals. To be more specific, this latest GPT-4 model accepts text and images inputs and produces text outputs, demonstrating its capability in conducting multimodal dialogues with humans. Within the realm of emotion recognition, the strategy of integrating information from multiple modalities is typically superior to relying solely on one single modality. Emotion, as a complex human phenomenon, is expressed through multiple channels, including text, speech, facial expressions, body language, and more. Instead of considering text only, jointly analysing multiple modalities can achieve a more holistic and accurate understanding of the emotional state being expressed. While LLMs provide a straightforward approach for estimating emotions based on text inputs, integrating these models with other modalities holds substantial promise. Future research directions may focus on investigating optimal approaches to fuse the multimodality information effectively without losing crucial emotional context.

Furthermore, other research efforts are being directed towards the development of large vision models~\cite{zhai2022scaling, liu2022swin}, large speech models~\cite{huang2023audiogpt, zhang2023google}, and multimodal large language models~\cite{yin2023survey}. Instead of utilising traditional models and gaining knowledge from small-scale emotional datasets, these large models hold immense potential to bring large gains in emotion recognition performance, in both single modality and multimodality. Incorporating these advanced models could potentially transform the future landscape of affective computing.









\section{Conclusion}
In this study, we provided a deep insight into the capability of Large language models (LLMs) in the domain of emotion recognition. We explored this by evaluating the performance of three leading LLMs across seven datasets, and compared the results against state-of-the-art works. The experimental outcomes have shown that the LLMs have shown superior performance in sentiment analysis and emotion recognition tasks, especially when identifying minority emotion categories. Unlike traditional specialised models, the LLMs benefit from being trained on expansive data volumes. This vast exposure equips them with impressive generalisation capabilities and paves the way for enhanced explainability in their predictions. Importantly, our findings also indicate that LLMs harness context to enhance emotion estimation. 
Furthermore, we have mapped out potential trajectories for the future of LLM-based emotion recognition systems. Our vision for these systems combines adaptability, multi-modality, robustness, privacy preservation, and compatibility with resource-constrained devices.


%



\section*{Acknowledgment}
The work leading to this research was partially supported by the National Science Foundation of China under Grant Number 62076092. 

\ifCLASSOPTIONcaptionsoff
  \newpage
\fi



%
\bibliographystyle{IEEEtran}
\bibliography{ref}




%
\vspace{-1cm}
\begin{IEEEbiography}[{\includegraphics[width=1in,height=1.25in,clip,keepaspectratio]{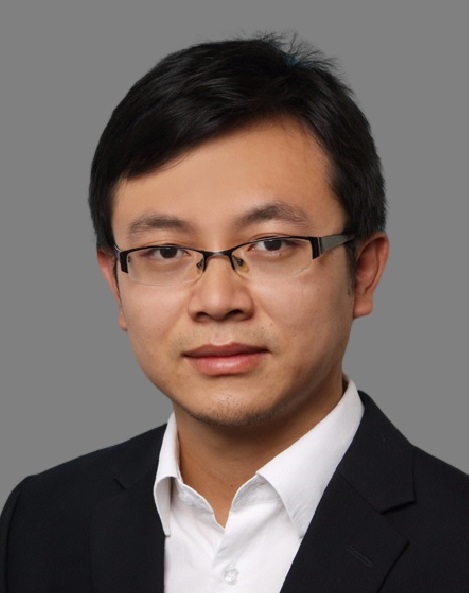}}]
{\bf Zixing Zhang} (M'15-SM'22) received his master degree in physical electronics from the Beijing University of Posts and Telecommunications (BUPT), China, in 2010, and his PhD degree in computer engineering from Technical University of Munich (TUM), Germany, in 2015. He is now a full professor at the College of Computer Science and Electronic Engineering, Hunan University, China. From 2017 to 2019, he was a research associate with the Department of Computing at the Imperial College London (ICL), UK. Before that, he was a postdoctoral researcher at the University of Passau, Germany. His research focuses on human-centred emotion and health computation. To date, he has authored more than 110 publications in peer-reviewed books, journals, and conference proceedings, leading to more than 5\,000 citations (h-index 40). He serves as an associate editor of the IEEE Transactions on Affective Computing and the Frontiers in Signal Processing, an editorial board member of the Nature Scientific Reports, and a guest editor of the IEEE Transactions on Emerging Topics in Computational Intelligence. 
\end{IEEEbiography}

\begin{IEEEbiography}[{\includegraphics[width=1in,height=1.25in,clip,keepaspectratio]{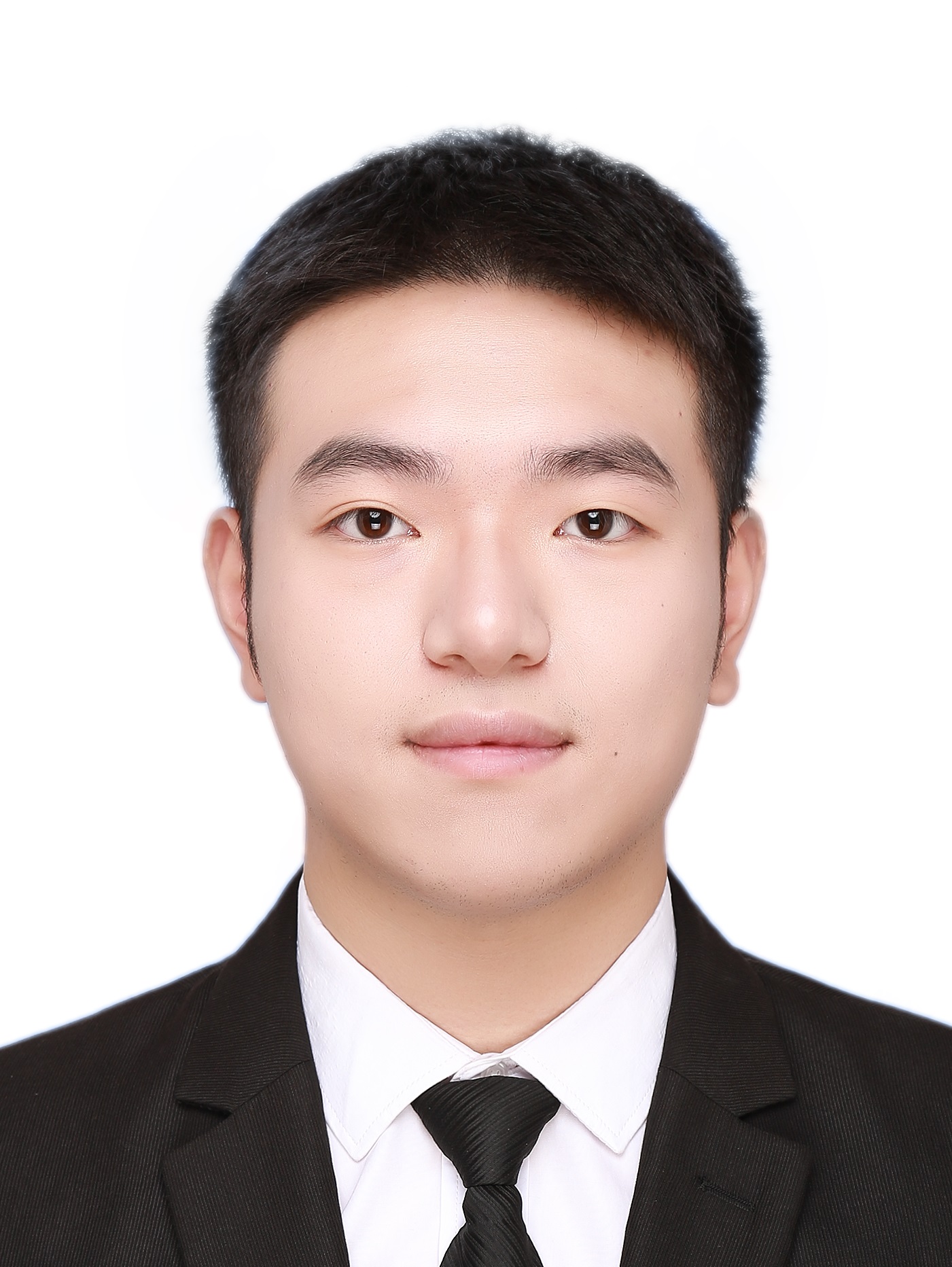}}] 
{\bf Liyizhe Peng} is currently a master student at the College of Computer Science and Electronic Engineering, Hunan University, China. His research interests include large language models, model adaptation, and emotion recognition.
\end{IEEEbiography}

\begin{IEEEbiography}[{\includegraphics[width=1in,height=1.25in,clip,keepaspectratio]{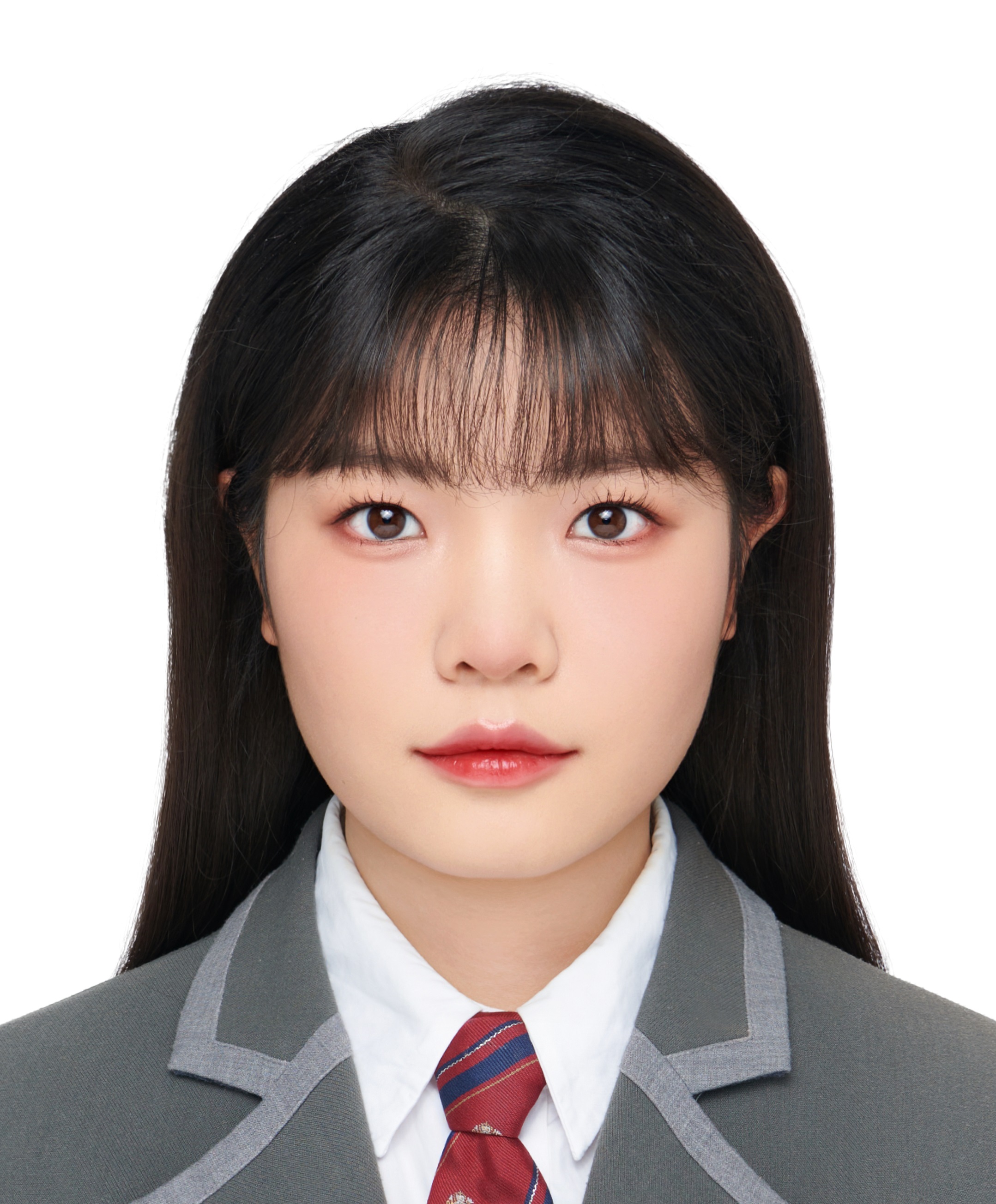}}] 
{\bf Pang Tao} is currently a master student at the College of Computer Science and Electronic Engineering, Hunan University, China. Her research topics relate to health-related audio signal processing.
\end{IEEEbiography}

\begin{IEEEbiography}[{\includegraphics[width=1in,height=1.25in,clip,keepaspectratio]{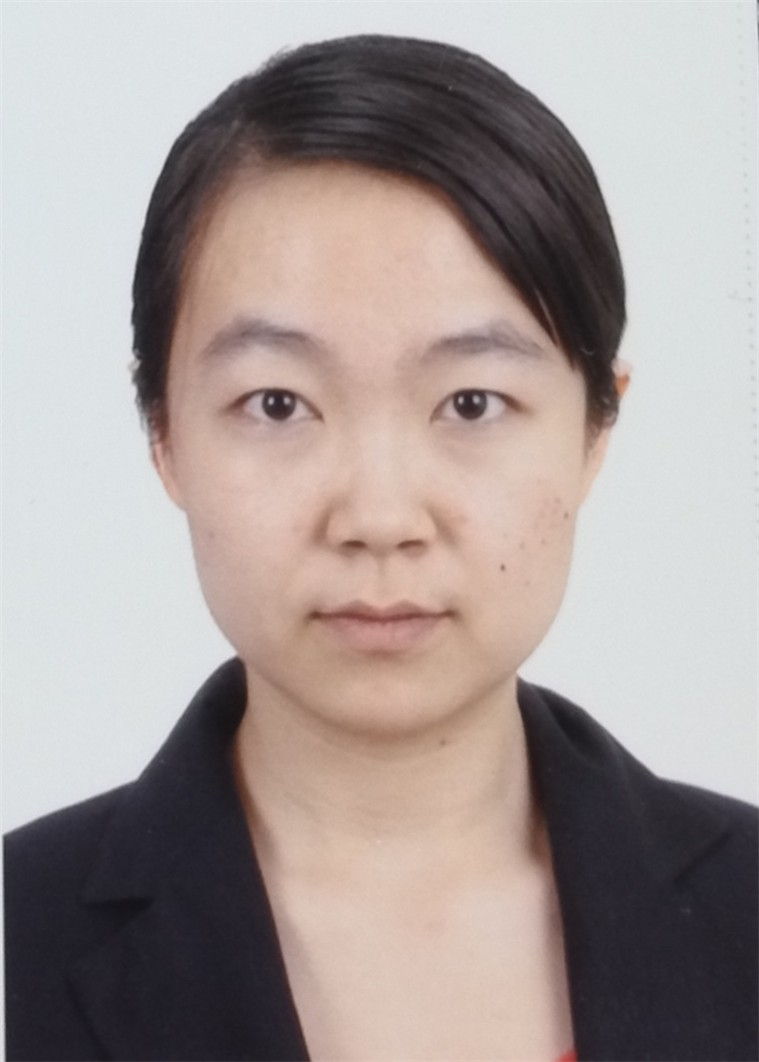}}]
{\bf Jing Han} (S'16-SM'23) received her bachelor degree (2011) in electronic and information engineering from Harbin Engineering University (HEU), China, her master degree (2014) from Nanyang Technological University, Singapore, and her PhD degree (2019) in Computer Science from University of Augsburg, Germany. Since 2019, she has been a postdoctoral researcher in the Department of Computer Science and Technology, University of Cambridge, UK. Her research interests are in affective computing and digital health. She (co)authored more than 60 publications in peer-reviewed journals and conference proceedings. 
She has served as a program committee member of the Audio/Visual Emotion Challenge and Workshop in 2018 and a technical program committee member of the Association for Computing Machinery (ACM) Multimedia since 2019, and is a leading guest editor of the IEEE Transactions on Emerging Topics in Computational Intelligence, and an associate editor of the IEEE Transactions on Affective Computing. She is a senior member of IEEE. 
\end{IEEEbiography}

\begin{IEEEbiography}[{\includegraphics[width=1in,height=1.25in,clip,keepaspectratio]{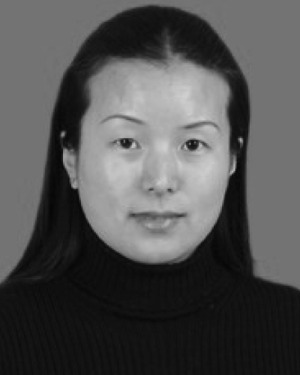}}] {\bf Huan Zhao} (M'10) received her Bachelor and Master degrees in computer application technology and her PhD degree in computer science and technology from Hunan University, in 1989, 2004, and 2010, respectively. She is currently a Professor at the College of Computer Science and Electronic Engineering, Hunan University. 
Her main research areas include speech information processing, natural language processing, and multimodal data analysis. She has led over 20 research projects funded by the National Key R\&D Program and the Natural Science Foundation of China. She has published over 100 papers in journals and conferences, such as ITCSVT, ICASSP, and the Journal of Communications, and has been granted eight patents.

\end{IEEEbiography}

\begin{IEEEbiography}[{\includegraphics[width=1in,height=1.25in,clip,keepaspectratio]{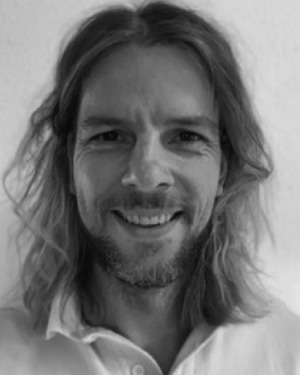}}] 
{\bf Bj\"orn~Schuller} (M'05-SM'15-F'18) received his diploma in 1999, his doctoral degree in 2006, and his habilitation and Adjunct Teaching Professorship in the subject area of signal processing and machine intelligence in 2012, all in electrical engineering and information technology from Technische Universit\"at Munchen (TUM), Germany. He is a tenured Full Professor heading the Chair of Embedded Intelligence for Health Care and Wellbeing, University of Augsburg, Germany, and a Professor of Artificial Intelligence heading GLAM, Department of Computing, at the Imperial College London in London, UK. Dr.~Schuller is the Field Chief Editor of Frontiers in Digital Health, former Editor in Chief of the IEEE Transactions on Affective Computing, President-emeritus and Fellow of the AAAC, Fellow of the IEEE, Golden Core Awardee of the IEEE Computer Society, Fellow of the ISCA, Fellow of the BCS, Fellow of the ELLIS, and Senior Member of the ACM. He (co-)authored 5 books and more than 1200 publications in peer-reviewed books, journals, and conference proceedings leading to more than 56\,k citations (h-index 106).
\end{IEEEbiography}







\end{document}